\documentclass[sn-basic]{sn-jnl-preprint}

\jyear{2022}

\newcounter{temp}

\graphicspath{{figures/}}

\def\ps@titlepage{
    \renewcommand{\@oddhead}{}
}
\def\ps@headings{
    \renewcommand{\@oddhead}{}
    \renewcommand{\@evenhead}{}
}

\newcommand*{\criteria}[1]{\textsc{#1}}

\newcommand*{\SC}[0]{SC\xspace}

\usepackage{xspace}
\usepackage{placeins}

\usepackage{amsthm}
\newtheorem*{problem*}{Problem}

\begin{document}

\title{Hitting the Target: Stopping Active Learning at the Cost-Based Optimum}

\author*[1]{\fnm{Zac} \sur{Pullar-Strecker}}\email{zpul156@aucklanduni.ac.nz}

\author[1]{\fnm{Katharina} \sur{Dost}}\email{kdos481@aucklanduni.ac.nz}
\author[2]{\fnm{Eibe} \sur{Frank}}\email{eibe@waikato.ac.nz}
\author[1]{\fnm{J\"{o}rg} \sur{Wicker}}\email{j.wicker@auckland.ac.nz}

\affil[1]{
    \orgname{University of Auckland}, 
    \orgaddress{\city{Auckland}, \country{New Zealand}}}
\affil[2]{
    \orgname{University of Waikato}, 
    \orgaddress{\city{Hamilton}, \country{New Zealand}}}

\abstract{
Active learning allows machine learning models to be trained using fewer labels while retaining similar performance to traditional supervised learning. An active learner selects the most informative data points, requests their labels, and retrains itself. While this approach is promising, it raises the question of how to determine when the model is `good enough' without the additional labels required for traditional evaluation.

Previously, different stopping criteria have been proposed aiming to identify the optimal stopping point. Yet, optimality can only be expressed as a domain-dependent trade-off between accuracy and the number of labels, and no criterion is superior in all applications. As a further complication, a comparison of criteria for a particular real-world application would require practitioners to collect additional labelled data they are aiming to avoid by using active learning in the first place.

This work enables practitioners to employ active learning by providing actionable recommendations for which stopping criteria are best for a given real-world scenario. We contribute the first large-scale comparison of stopping criteria for pool-based active learning, using a cost measure to quantify the accuracy/label trade-off, public implementations of all stopping criteria we evaluate, and an open-source framework for evaluating stopping criteria. Our research enables practitioners to substantially reduce labelling costs by utilizing the stopping criterion which best suits their domain.
}


\maketitle

\section{Introduction}
\label{sec:introduction}


Labelling a large enough sample of data for supervised machine learning (ML) can be prohibitively expensive. This holds back the deployment of learning algorithms in practical applications where the cost of each label is high. For example, developing a model to determine if cancer is present in a mammogram traditionally requires a large number of images manually labelled by an expert. The problem is exacerbated if labelling requires not only expensive but also time-consuming experiments as in functional genomics \citep{king_functional_2004}. Hence, our goal is to train a model that makes as few mistakes as possible, using as few labels as possible.

\begin{figure}
    \centering
    \includegraphics[width=0.7\textwidth]{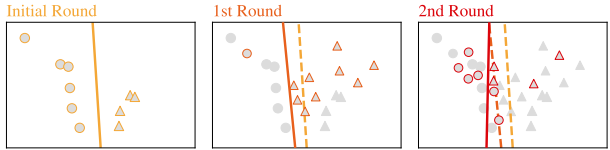}
    \caption{A \textit{support vector machine} is initially trained on $10$ points in a toy dataset (yellow). In the first round of AL, $10$ points are selected by the query strategy and then the classifier is retrained on $20$ labelled points (orange). In the second round, a further $10$ points are selected, and the classifier is retrained on $30$ labelled points (red)}
    \label{fig:al_2d}
\end{figure}

\textit{Active learning} (AL) is a strategy that seeks to address this problem by training ML models using fewer labels than traditional random sampling. In this method, the model is first trained on a small set (pool) of labelled data. 
Then, an AL strategy selects a small number of unlabelled samples it expects to learn the most from if they are labelled. Labels are then obtained by, for example, carrying out experiments in a lab or consulting a domain expert \citep{settles_active_2012}.
The newly labelled samples are added to the training set, the model is re-trained, and the process is repeated. A snapshot of this process is shown in Fig. \ref{fig:al_2d}. 

Active learning is based on the premise that labels are expensive, and that the number collected should be minimised. A key question is when to stop the AL process, since the active learner would be no better than the traditional passive one otherwise. Either training is stopped too early and the obtained model is not fit for purpose, or too late, incurring unnecessary cost without improvement of the model. The na\"ive approach is to collect extra labels for a validation set to evaluate the performance of the classifier. Unfortunately, this comes at a substantial cost and defeats the point of using AL in the first place \citep{beatty_use_2019}. Hence, to make AL useful in practical applications, there is a need to find a \textit{stopping criterion} to determine the best time to stop AL
(to \textit{hit the target})
based on the information already available: the labelled training pool, the unlabelled pool, and the classifier itself.



The inherent trade-off between the number of instances labelled and the accuracy of the resulting classifier makes choosing stopping criteria (SC) challenging. Prior work has introduced a large number of \SC, each with their own hyperparameters, without providing any method or guidelines to determine which is appropriate for a given scenario. 
This leaves a practitioner no other choice than either blindly trusting a stopping criterion, or collecting additional labels to carry out a small benchmark comparison, which defeats the point of AL.
To address this issue, we introduce a method that uses a cost measure to connect the trade-off to intuitive parameters such as label cost and model lifetime. We use our measure to perform a large-scale comparison between prior \SC across a variety of datasets and models to identify which criterion is best for a particular task.
We summarise our main contributions as follows:
\begin{itemize}
    \item We introduce a cost measure-based approach for \SC selection which is crucial for practitioners to pick an appropriate criterion for their use-case.
    \item Our large-scale comparison across three learning models and nine datasets using our cost measure maps $13$ \SC to the most appropriate domain.
    \item We provide the first open source implementation of all \SC we evaluate making them available to practitioners and supporting researchers to compare and evaluate \SC more easily in a unified framework.
    \item Lastly, we discuss a set of promising directions for future development that arise from our investigations and might guide future researchers to performant \SC and robust comparisons.
\end{itemize}


The remainder of this paper is organised as follows:
Section \ref{sec:problem} provides a formal definition of AL and \SC, then Section \ref{sec:related_work} reviews prior \SC. Section \ref{sec:cost_measure} introduces our cost measure. Experimental results and arising research opportunities are discussed in Sections \ref{sec:experiments} and \ref{sec:limitations}, respectively. Finally, Section \ref{sec:conclusion} concludes.

\section{Problem Statement}
\label{sec:problem}
Active learning describes a strategy that, given a small number of labelled points and a classifier trained on them, selects examples that should be labelled in order to best improve the classifier's performance. Active learning does not specify how the labels are obtained, they could come from a human, robot, or other model \citep{settles_active_2012}. We consider the labeller a black box which we name the \textit{oracle}.


A number of different settings for AL have been proposed. These differ in how the query strategy obtains points to be labelled.
The two most popular settings are \textit{(batch-mode) pool-based AL}, and \textit{query-synthesis AL} \citep{settles_active_2012}. In pool-based AL, points are selected from a pre-existing pool of unlabelled points. Batch-mode pool-based AL is an extension to this method where a batch of points is selected at a time, primarily to reduce the number of times the model needs to be retrained.
Query-synthesis AL does not require a pool of unlabelled instances and instead generates points de novo. This setting is difficult to use in practise as most query strategies generate points that are not easily interpretable by a human \citep{baum_query_1992}. 
Most of the query-synthesis research to date is purely theoretical. In order to use query-synthesis AL in practice, the point generator needs to be strongly restricted to produce only meaningful points within the domain space (for example, only feasible compounds in the chemical compound space). However, if such a generator exists it can be exploited to obtain an unlabelled sample enabling pool-based AL. Hence, in this paper, we focus on pool-based AL, which is formally defined below.

\begin{problem*}[Batch-Mode Pool-Based AL]
Let $\mathcal{L}_0$ be a small labelled dataset that contains at least one example of each class, and $\mathcal{U}$ be the unlabelled dataset. 
First, a classifier $\mathbf{C}_0$ is trained on $\mathcal{L}_0$, and a batch-mode pool-based AL strategy $Q$  selects the $k$ examples $x_1,\dots,x_k \in \mathcal{U}$ it deems the most informative from the pool. 
Second, these samples are labelled by the oracle $\mathbf{O}$, added to the labelled pool $\mathcal{L}_0$ to obtain $\mathcal{L}_1 = \mathcal{L}_0 \cup_{i=1}^k \{(x_i, \mathbf{O}(x_i))\}$, and removed from the unlabelled pool $\mathcal{U}$. Third, a classifier $\mathbf{C}_1$ is trained on $\mathcal{L}_1$. This process is repeated until we decide to stop or the unlabelled pool is empty.
\end{problem*}


Deciding when to stop AL is determined by \SC. To aid explanation, we split each criterion into two parts. First, a \textit{metric} $m\colon \mathcal{U}, \mathcal{L}, \mathbf{C}\rightarrow \mathbb{R}$, which takes the classifier $\mathbf{C}$ as well as the labelled and unlabelled pools ($\mathcal{L}$ and $\mathcal{U}$), and returns a real number.
Second, a \textit{condition} $c\colon \mathbb{R}^i\rightarrow \{0,1\}$, which, in each iteration $i$, takes this number (and optionally the metric values from the previous $i-1$ iterations) and returns a Boolean indicating whether the active learner should stop ($1$) or not ($0$). 
By splitting each stopping criterion into a metric and a condition, we gain the ability to analyze them separately. A good metric should exhibit an obvious change (e.g., a critical point) near the point at which the classifier's accuracy saturates. A good condition should reliably and accurately locate this point and stop the AL process.

\section{Related Work}
\label{sec:related_work}


\setcounter{temp}{\value{footnote}}
\setcounter{footnote}{0}
\begin{table}[tbp]
    \begin{center}
        \begin{minipage}{\textwidth}
            \renewcommand{\thefootnote}{\alph{footnote}}
            \caption{The \SC evaluated in this paper. `Metric' and `Condition' describe each criterion's design. `Models' indicates the learning algorithms supported by each criterion.}\label{tab:criteria}%
            \centering
            \begin{tabular}{p{4cm}|p{3.5cm}|l|l}
                \toprule
                Criterion & Metric & Condition & Models \\
                \midrule
                \criteria{Max-confidence} & Selected uncertainty & Threshold & Prob \\
                \criteria{Entropy-MCS} & Selected uncertainty & Threshold & Prob. \\
                \criteria{MES} & Expected error & Threshold & Prob. \\
                \criteria{OracleAcc-MCS} & Selected accuracy & Threshold & All \\
                \criteria{Classification change} & Classifier agreement\footnotemark & Threshold & All \\
                \criteria{Overall uncertainty} & Uncertainty & Threshold & Prob. \\
                \criteria{Performance convergence} & Lewis estimate & Stationary & Prob. \\
                \criteria{Uncertainty convergence} & Selected uncertainty & Stationary & Prob. \\
                \criteria{Contradictory information} & Contradictory information & Maxima  & Prob. \\
                \criteria{Stabilizing predictions} & Classifier agreement & Threshold & All \\
                \criteria{VM} & Uncertainty variance & Maxima & Prob. \\
                \criteria{EVM} & Uncertainty variance & Maxima & Prob. \\
                \criteria{SSNCut} & Spectral agreement & Stationary & SVMs \\
                \bottomrule
            \end{tabular}    
            \footnotetext[a]{This is distinct from, though similar to, the metric used by \criteria{Stabilizing predictions}.}
        \end{minipage}
    \end{center}
\end{table}
\setcounter{footnote}{\value{temp}}

Several approaches have been introduced that can be applied to the batch-mode pool-based AL problem, but, to the best of our knowledge, no thorough comparison of the \SC has been provided. This section provides an overview of previously proposed \SC detailing their conditions and metrics. A summary of this information can also be found in Table \ref{tab:criteria}.

\criteria{Stochastic search normalized cut} (\criteria{SSNCut}) trains a spectral clustering model on the unlabelled pool \citep{fu_low_2015}. The metric is the proportion of disagreements between the cluster and the classifier labels. If this proportion has not reached a new minimum for $10$ iterations (i.e., it has stopped decreasing) the classifier is retrained with only the samples that were available $10$ iterations ago; this is the condition. \criteria{SSNCut} is designed for binary classification tasks using support vector machines (SVMs), so while it is not general the unusual approach makes it an interesting criterion to evaluate.

\criteria{Entropy-MCS}'s metric is the maximum of the classifier's entropy in the estimated class probability distribution of instances in the unlabelled pool \citep{zhu_learning_2008}. The condition stops AL if the metric is less than or equal to a threshold. The authors suggested three values for the threshold: $0.01, 0.001, 0.0001$.

\criteria{OracleAcc-MCS} evaluates the accuracy of the classifier on samples that have just been labelled by the oracle, but not yet been trained on \citep{zhu_learning_2008}. It relies on the accuracy of instances selected in a single round so it can only be used in the batch-mode AL setting. 
The authors mention the batch size should not be too small but otherwise provide no additional guidance.
The accuracy is the metric, and the condition is a threshold with suggested values of $0.9$, and $1.0$.

\criteria{Minimum expected error} (\criteria{MES}) uses the expected error of the current classifier as the metric and a threshold as the condition \citep{zhu_learning_2008}. The metric is calculated as follows: 
 $\textrm{Error}(\mathbf{C}_i)=\lvert\mathcal{U}\rvert ^{-1}\sum_{x\in\mathcal{U}}\left(1-\max_{y\in \mathcal{Y}}\mathbb{P}(y\mid x)\right)$
where $\mathcal{Y}$ is the set of possible labels and $\mathbb{P}(y\mid x)$ is the posterior probability distribution of the current classifier $\mathbf{C}_i$. This was originally introduced as a loss function for minimum-error classification which aims to maximise the probability of the correct class \citep{duda_pattern_2000}. 
The suggested values for the threshold were $0.01, 0.001, 0.0001$. 

\criteria{Contradictory information} aims to estimate the amount of contradictory information added in each round \citep{vlachos_stopping_2008}. In the original paper, the contradictory information is based upon the decision margin of an SVM. However, we use the confidence of the model to extend it to any probabilistic model.
The metric is calculated after selecting labels but before retraining the new classifier. The old classifier is used to predict probabilities for each of the queried instances, and then the mean probability of incorrectly predicted instances is used as the metric.
\citeauthor{vlachos_stopping_2008} does not evaluate a condition, but suggests that if the amount of contradictory information drops for three consecutive rounds, this may correlate with a good stopping point. We use this condition in our evaluation.

\criteria{Stabilizing predictions} records the change of the classifiers' predictions on a \textit{stop set} \citep{bloodgood_method_2009}. The stop set is a subset of the unlabelled pool whose instances may be used in training.
Each of the last $k$ classifiers are used to predict the labels on the stop set. Then, the Kappa statistic is evaluated on all pairs of predictions and averaged to obtain the metric \citep{cohen_coefficient_1960}. The condition is a threshold, such that the AL stops once it is reached or exceeded; the authors suggest $0.99$.

\criteria{Variance model} (\criteria{VM}) stops when the variance of the classifier's uncertainty on the unlabelled pool decreases in $n$ sequential iterations with the authors suggesting $n=2$ \citep{ghayoomi_using_2010}. \citeauthor{ghayoomi_using_2010} also introduces \criteria{Extended Variance Model} (\criteria{EVM}), which uses the same metric, but extends the condition to require that the metric decreases for $n$ sequential iterations by a value of at least $m$. The suggested value for $m$ is $0.001$.

\criteria{Performance convergence} uses the Lewis estimate of F-score extended to multi-class classification and finds a derivative via differences over a sliding window of the last $k$ rounds of AL \citep{laws_stopping_2008}. The Lewis estimate approximates F-score by estimating the true positive, false positive, and false negative proportions based on the classifier's uncertainty. This method was tested using both the mean and the median over the sliding window. The condition requires that (i) the metric is at a new maximum, (ii) the derivative is positive, and (iii) the derivative is less than a threshold $\epsilon$. \citeauthor{laws_stopping_2008} suggest $k=10$, $\epsilon=5\times10^{-5}$, and aggregating using the mean.

\criteria{Uncertainty convergence} uses the entropy of the last instance selected by the query strategy \citep{laws_stopping_2008}. Because we evaluate \SC in a batch-mode setting, we extend this metric by taking the minimum uncertainty across the selected instances. As in \criteria{Performance convergence}, a derivative is found via a sliding window over the last $k$ rounds. The authors define the maximum uncertainty as $0$ and the minimum uncertainty as $1$, we swap this to the conventional definition of $1$ as the maximum uncertainty and $0$ as the minimum uncertainty. This method was only tested using the median over the sliding window in \citep{laws_stopping_2008}. The condition is the same as in \criteria{Performance convergence}, except we flip the first condition (the metric is at a new minimum) to match the conventional uncertainty definition.

\criteria{Max confidence} also uses the classifier's uncertainty on the last instance selected by the query strategy \citep{zhu_active_2007}. As for \criteria{Uncertainty convergence}, we extend the uncertainty on selected instances by taking the minimum. The condition to determine the stopping point is a threshold, with the authors' suggesting $0.001$.

\criteria{Classification change} stops if there are no changes in the predictions of the classifier on the unlabelled samples between two rounds \citep{zhu_multi-criteria-based_2008}. This is the first criterion discussed that does not have a separation between the metric and condition. For the purposes of visualisation we redefine this method by first calculating a metric, the proportion of equal label predictions between the two classifiers, and a condition consisting of a threshold at $1.0$. This is functionally identical to the original method.

\criteria{Overall uncertainty} uses the uncertainty of the classifier on the unlabelled pool as a metric \citep{zhu_multi-criteria-based_2008}. The condition is a threshold for which \citeauthor{zhu_multi-criteria-based_2008} suggest $0.01$.

A number of other \SC have been proposed that either require specific models or only apply to particular domains. We list them here for completeness, but omit them from our evaluation as they are not generally applicable. Model-specific \SC include those that require ensembles or a Bayesian model \citep{ishibashi_stopping_2020, ishibashi_stopping_2021}. Domain-specific \SC include work on drug target prediction, automated screening in systematic reviews, and a larger body of work focused on structural reliability analysis \citep{callaghan_statistical_2020, moustapha_generalized_2021, yi_efficient_2020}. 

The performance of \SC in a particular domain can be compared using a cost measure such as the one introduced by \cite{dimitrakakis_cost-minimising_2008}. Their cost measure is defined as:
\begin{equation*}
    \mathbb{E}[C_\gamma\mid f(t)]=\mathbb{E}[R\mid f(t)]+\gamma t,
\end{equation*}
where $f(t)$ is the hypothesis obtained after labelling $t$ examples, $\mathbb{E}[R \mid f(t)]$ is the generalisation error associated with $f(t)$, and $\gamma\in[0,\infty]$ is a hyperparameter describing the relative cost of classification error and obtaining labels. The authors use this cost measure to introduce a stopping condition, \criteria{OBSV}, that takes the cost hyperparameter $\gamma$ into account. This method, however, relies on passive (random) selection to obtain independently and identically distributed samples and determine a stopping point. Assuming we are in a `good' scenario for AL where the query strategy is capable of selecting informative examples these passively selected instances come at a higher cost because of the lower information the model is expected to learn from them. Additionally, this query behaviour makes it difficult to evaluate fairly against traditional \SC. For these reasons we do not evaluate \criteria{OBSV} in our study.

\section{Quantifying the Stopping Criteria Trade-off}
\label{sec:cost_measure}

The primary motivation for AL is to reduce the cost of labelling data, however, there is no single criterion that provides satisfactory results in \textit{all} applications. Depending on the domain, a practitioner has to consider several costs associated with labelling data and misclassifications of the trained model. In order for AL to provide value, these costs need to be considered when choosing a stopping criterion. Aiming to quantify the inherent AL trade-off, we express a combined cost that can be easily calculated based on the use-case as follows:
\begin{equation}
    \label{eq:cost}
    C(a, j, l, m, n)=(1-a)mn+jl,
\end{equation}
where $a$ and $j$ represent the accuracy and number of labels used by the \SC at the stopping point respectively, $l$ is the cost to obtain a single label, $m$ is the cost per misclassification, and $n$ is the expected number of predictions over the lifetime of the model. 
The misclassification cost $m$ can be further broken down if there are differing costs for incorrect predictions of each class given an approximation of the relative rates of misclassification.

While building upon the cost measure proposed by \cite{dimitrakakis_cost-minimising_2008}, our formulation enhances interpretability and usability in practice due to two key aspects.
First, the parameters in Eq. \ref{eq:cost} have practical meaning: we allow practitioners to use the directly measurable quantities misclassification cost, label cost, and model lifetime instead of an arbitrary parameter $\gamma$. 
Second, while their measure results in a uninterpretable unit-free numerical, our measure can be expressed as a currency, enabling easy selection and communication of a suitable trade-off.

We use results from our large-scale comparison to find the mean accuracy ($a$) and number of labels used ($j$). Then, given values for $n$, $m$, and $l$, the best criterion is the one that minimizes the cost $C$.

\section{Evaluation and Comparison}
\label{sec:experiments}

In order to apply AL effectively it is necessary to choose an appropriate stopping criterion. Yet, the lack of labels makes picking a criterion difficult as we do not have a hold-out set on which to evaluate \SC. We performed a large comparison of \SC across a variety of datasets and learning algorithms to answer the question of which criterion is best for a given domain. Because this is a multi-objective problem (i.e., we aim to minimize the number of obtained labels while maximizing accuracy), there is no objectively best criterion.
Sec. \ref{sec:experiments:qualitative} provides a qualitative comparison. Then, in Sec. \ref{sec:experiments:cost} we identify the best criterion for given cost ranges. Next, in Sec. \ref{sec:experiments:examples} we use our cost measure to identify the best \SC in two example scenarios. Sec. \ref{sec:experiments:models} describes the behaviour of \SC under random forests and NNs. The supplementary material and open source implementations of the \SC we evaluated are available on GitHub\footnote{The code and supplementary material is available at \href{github.com/zacps/al-stopping-conditions}{https://github.com/zacps/al-stopping-conditions}}.

\subsection{Setup}
\label{sec:experiments:setup}

To enable reproducibility, we detail our experimental setup including learning algorithms, query strategy, datasets, hyperparameters, and AL set sizes.

The models used are linear support vector machines, random forests, and a single hidden layer neural network. All model parameters are left at their default values. This is because during AL our only labelled data at the start of the process is the initial labelled set of $10$ instances, which is too small to perform traditional hyper-parameter tuning \citep{oliver_realistic_2019}.

Some of the \SC we evaluate require an estimate of the classifier's confidence to compute their metric. For the classifiers we use, SVMs, random forests, and NN, the posterior probabilities are not necessarily properly calibrated probabilities and might over or under predict the true value. However, all of the \SC we evaluate were designed on classifiers utilising these approximations, and the estimates are sufficient to point the active learner towards informative regions.
To estimate the confidence in each classifier, we use the following standard methods: SVM's confidence is proportional to the distance from the instance to the separating hyperplane. In random forests, it is the mean maximum predicted class probabilities of the trees in the forest, where the class probability of a single tree is the fraction of samples of the predicted class in the leaf. The confidence in NNs is the output of the softmax function on the last hidden layer for the predicted class.

As the query strategy we use \texttt{modAL}'s implementation of Cardoso's ranked-batch-mode uncertainty query-strategy with a batch size of $10$ \citep{cardoso_ranked_2017,danka_modal_2018}. 
We choose this strategy because it takes into account both aleatoric and epistemic uncertainty and performs well in existing empirical comparisons \citep{gissin_discriminative_2019}.
The strategy picks instances for which the model's prediction has low confidence, while also using a distance metric to ensure diversity among the selection.

Table \ref{table:datasets} shows the nine datasets we use in our evaluation. These were selected based on the following criteria: all datasets contain at least $3,000$ instances, have no missing values, and have a high AL `potential'. We define potential as the difference between the accuracy of the model after labelling $50\%$ of the instances (all bar the test set), and after labelling only $10$ instances. The potential describes the extent to which a dataset can benefit from AL. Other datasets either have a low final accuracy (i.e., the model is a poor fit) or a high initial accuracy (i.e., AL is unnecessary).

\setcounter{temp}{\value{footnote}}
\setcounter{footnote}{0}
\begin{table}[tbp]
    \begin{center}
        \begin{minipage}{290pt}
            \renewcommand{\thefootnote}{\alph{footnote}}
            \caption{Datasets used in experiments. AL Potential is the difference in accuracy between training on $50\%$ of the dataset and training on $10$ labels}\label{table:datasets}
            \begin{tabular}{lrrrll}
                \toprule
                 Dataset      &   Instances &   Classes &   Features & AL Potential & Domain   \\
                \midrule
                 sensorless\footnotemark[1]   &       $58\,509$ &        $11$ &         $48$ & $0.73\pm0.03$ & General  \\
                 webkb\footnotemark[2]        &        $4\,199$ &         $4$ &      $22\,981$ & $0.50\pm0.01$ & NLP      \\
                 swarm\footnotemark[1]        &       $24\,015$ &         $2$ &       $2\,400$ & $0.45\pm0.01$ & General  \\
                 smartphone\footnotemark[3]   &       $10\,927$ &        $12$ &        $561$ & $0.42\pm0.01$ & General  \\
                 rcv1\footnotemark[4]         &       $58\,509$\rlap{\footnotemark[7]} &         $2$&      $47\,236$ & $0.40\pm0.01$ & NLP      \\
                 splice\footnotemark[1]       &        $3\,190$ &         $3$ &        $287$ & $0.38\pm 0.02$ & General  \\
                 avila\footnotemark[5]        &       $20\,867$ &        $12$ &         $10$ & $0.37\pm0.01$ & General  \\
                 anuran\footnotemark[1]       &        $7\,195$ &        $10$ &         $22$ & $0.28\pm 0.02$ & General  \\
                 spamassassin\footnotemark[6] &        $6\,051$ &         $2$ &      $50\,196$ & $0.24\pm 0.01$ & NLP      \\
                \bottomrule
            \end{tabular}
            \footnotemark[1]\cite{dua_uci_2017} \ \footnotemark[2]\url{cs.cmu.edu/~webkb} \ \footnotemark[3]\cite{anguita_public_2013} \ \footnotemark[4]\cite{lewis_rcv1_2004} \ \footnotemark[5]\cite{destefano_reliable_2018} \ \footnotemark[6]\url{spamassassin.apache.org/old/publiccorpus/readme.html} \ \footnotemark[7]Indicates a subsample from a larger dataset.
        \end{minipage}
    \end{center}
\end{table}
\setcounter{footnote}{\value{temp}}

Hyperparameters for \SC were set to the values suggested by the authors (as listed in Sec. \ref{sec:related_work}) in cases where they suggested only a single option. In the cases where multiple values were suggested (i.e., for \criteria{Entropy-MCS}, \criteria{MES}, \criteria{OracleAcc-MCS}), we set the threshold to be the value that is the most likely to be reached ($0.01$, $0.01$, and $0.9$ respectively). In preliminary experiments, these \SC were not close to meeting the stricter thresholds and often did not stop at all, so this gives them the best chance of performing competitively.

To create the initial labelled set we selected a random instance from each class, then randomly added instances until there were at least $10$ in the initial set.
We pick this value to balance maximising the number of labels that are actively learnt without the variance of the initial labelled set being too high.
While not possible in practise, this ensures a constant initial set size for each dataset and is unlikely to affect the performance of the \SC. Common choices in prior work are $1$ per class, $100$, and $1\%$ \citep{gissin_discriminative_2019}.

To reduce the computational cost, we sub-sample $1,000$ unlabelled instances to use for query strategy and metric evaluation during the AL rounds. Sub-sampling the unlabelled pool is common practise, with subset sizes from $500$ to a few thousand instances \citep{zhu_learning_2008, fu_low_2015}. As an additional measure to reduce computational expenditure, we use only $58,509$ instances from rcv1 (the size of the next smallest dataset sensorless).

We continued AL until there were $500$ instances left in the unlabelled pool. These instances were not used because the \SC might stop solely due to the altered distribution of the unlabelled pool caused by the biased sampling. In this case, the \SC would not be detecting any real change in the classifier's performance. The number of reserve instances was chosen to balance avoiding this case while observing the \SC for as long as possible. In all experiments, $50\%$ of each dataset was held out as a test set. We repeat all experiments $30$ times with randomised splits to produce reliable results. 
The initial labelled set, queried points, and classifiers are identical for all stopping criteria on a given random split.

\subsection{Qualitative Comparison}
\label{sec:experiments:qualitative}

\begin{table}[tbp]
    \begin{center}
        \begin{minipage}{\textwidth}
            \caption{Summary of the behaviour of \SC. ``Behaviour'' indicates our qualitative evaluation of when they tend to stop, ``Dominates'' describes the relative label costs the criteria are best for according to our cost analysis, ``Stops'' indicates the fraction of the time the criteria stop in our tests, and ``Accuracy correlation'' shows the correlation between each criterion's metric and the classifier's accuracy}
            \label{tab:criteria_results}
            \begin{tabular}{l|l|l|r|r}
                \toprule
                Criterion & Behaviour & Dominates & Stops & \multicolumn{1}{p{1.48cm}}{Accuracy Correlation} \\
                \midrule
                \criteria{Max-confidence} & Inconsistent & - & $68\%$ & $0.55\pm0.06$ \\
                \criteria{Entropy-MCS} & Inconsistent & - & $11\%$ & $0.47\pm0.04$ \\
                \criteria{MES} & Late & - & $83\%$ & $0.69\pm0.06$ \\
                \criteria{OracleAcc-MCS} & Inconsistent & - & $89\%$ & $0.53\pm0.05$ \\
                \criteria{Classification change} & Aggressive & Cheap labels & $100\%$ & $0.73\pm0.04$ \\
                \criteria{Overall uncertainty} & Late & - & $83\%$ & $0.69\pm0.06$ \\
                \criteria{Performance convergence} & Late & - & $75\%$ & $0.69\pm0.06$ \\
                \criteria{Uncertainty convergence} & Inconsistent & - & $22\%$ & $0.50\pm0.04$ \\
                \criteria{Contradictory information} & Early & Expensive labels & $100\%$ & $0.08\pm0.04$ \\
                \criteria{Stabilizing predictions} & Aggressive & Balanced costs & $100\%$ & $0.85\pm0.03$ \\
                \criteria{VM} & Early & Expensive labels & $100\%$ & $0.27\pm0.08$ \\
                \criteria{EVM} & Early & - & $89\%$ & $0.27\pm0.08$ \\
                \criteria{SSNCut} & Early & Expensive labels & $100\%$ & $0.17\pm0.04$ \\
                \bottomrule
            \end{tabular}
        \end{minipage}
    \end{center}
\end{table}

Our results are summarised in Table \ref{tab:criteria_results}. The columns labelled ``Behaviour'', ``Stops'', and ``Correlation'' are discussed below, and the  column labelled ``Dominates'' is discussed in Sec. \ref{sec:experiments:cost}.

\begin{figure}
    \centering
    \includegraphics[width=\textwidth]{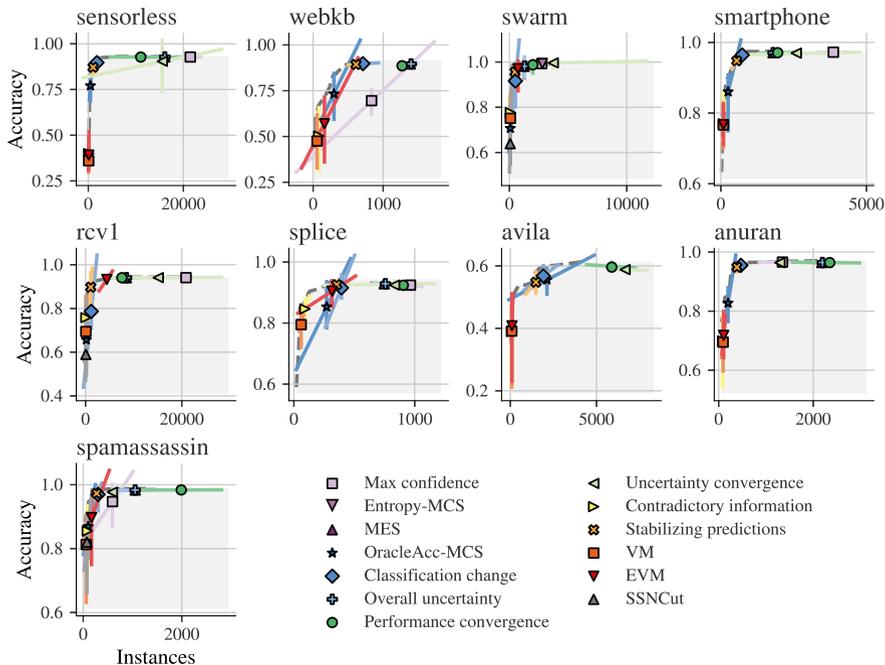}
    \caption{The accuracy and number of instances used by \SC when evaluated for SVMs. The dashed grey line shows the Pareto frontier. Error bars were computed from $2.5\%$ and $97.5\%$ percentiles and transformed by PCA}
    \label{fig:pareto_svm}
\end{figure}

In our primary experiments, we evaluate the performance of the 12 competing \SC discussed in Section \ref{sec:related_work} on SVMs. Figure \ref{fig:pareto_svm} shows the number of labels and the accuracy of the classifier when each criterion indicated that AL should stop. The dashed grey line shows the \textit{Pareto frontier}. If a run of a criterion is on the Pareto frontier, it means that no other criterion run has \textit{both} better accuracy and uses fewer labels. 
We begin our discussion by qualitatively dividing the \SC into four groups: early, aggressive, late, and inconsistent.

\begin{figure}
    \centering
    \includegraphics[width=0.9\textwidth]{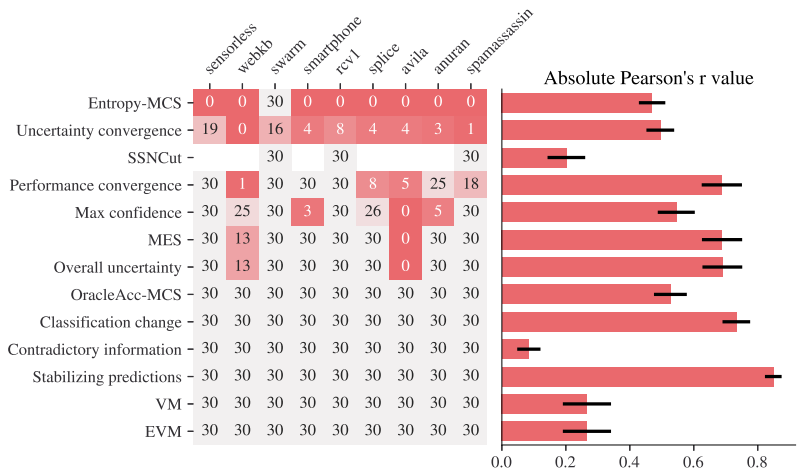}
    \caption{Left: the number of times each criterion stopped on each dataset using SVMs, out of $30$ splits. Right: the Pearson correlation coefficient between each \SC's metric and the classifier's accuracy. The black line shows the standard error of the mean across datasets}
    \label{fig:failed_to_stop}
\end{figure}

The \textit{early} \SC (\criteria{VM}/\criteria{EVM}, \criteria{SSNCut}, and \criteria{Contradictory Information}) all stop while the classifier's accuracy is rapidly increasing. On most of the datasets we evaluate, the accuracy at this point is significantly below the maximum accuracy achieved by the classifier. As this is a multi-objective problem, these \SC cannot be discarded simply because of their low accuracy at the stopping point. Yet, we suspect that they stop too early to be desirable to most practitioners.
These methods have metrics which have a low correlation with the classifier's accuracy. In all cases the absolute value of Pearson's correlation coefficient is less than $0.4$, which is shown in Fig. \ref{fig:failed_to_stop}.

The second group of \SC (\criteria{Stabilizing Predictions}, \criteria{Classification Change}) stop \textit{aggressively}, close to the point at which the classifier's performance is saturated. Stopping near the saturation point yields the highest accuracy per label, and so stopping near this point is desirable for a large range of label and misclassification costs. 
Both of these \SC use the change in the classifier's predictions between rounds to identify a stopping point, indicating this approach is a good choice for detecting the saturation point.

Three \SC, \criteria{Overall uncertainty}, \criteria{Performance convergence} and \criteria{MES}, stop \textit{late}. At this point, the classifier's performance has saturated and little can be gained by continuing to train. In all but exceptionally high misclassification cost scenarios, these \SC will use more labels than necessary to train a useful model.

Lastly, we consider \SC that stop \textit{inconsistently}, if ever. These are \criteria{OracleAcc-MCS}, \criteria{Entropy-MCS}, \criteria{Uncertainty convergence}, and \criteria{Max confidence}. These \SC either fail to stop at all during our evaluation, or stop erratically. If they did not stop it is possible that they may eventually stop, but by that point they would have used far more labels than necessary to train a performant classifier. Hence, it is unlikely they will be useful to practitioners. In the case of \criteria{Uncertainty convergence} and \criteria{Max confidence}, the metric has some correlation with the classifier's accuracy (see Fig. \ref{fig:failed_to_stop}). However, because the metric has a high variance, the condition used to determine the stopping point is fragile. \criteria{OracleAcc-MCS} and \criteria{Entropy-MCS} both have metrics that vary between datasets, so finding a criterion that stops consistently may not be possible.

Out of the $13$ \SC we evaluate, six stop in all of our tests. \SC that did not stop are undesirable as they are likely to use an excessive number of labels, or never halt at all. Figure \ref{fig:failed_to_stop} shows the number of splits on each dataset that each criterion stopped on, as well as the mean Pearson correlation coefficient between the classifier's accuracy and \SC's metrics by dataset. \criteria{Entropy-MCS} and \criteria{Uncertainty convergence} frequently fail to stop, making them difficult to recommend as they are not likely to stop at a consistent or predictable time. \criteria{SSNCut} was only evaluated on binary datasets.

\subsection{Cost-based Analysis}
\label{sec:experiments:cost}

\begin{figure}
    \centering
    \includegraphics[width=\textwidth]{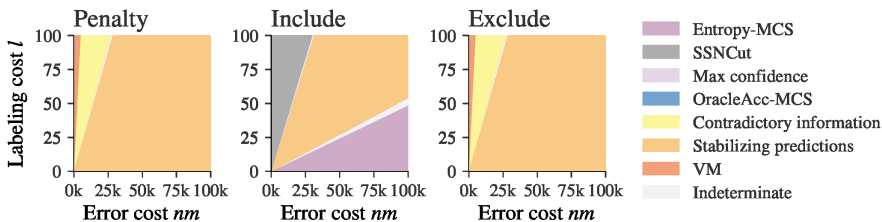}
    \caption{The regions in which each criterion is cost-optimal for SVMs. The three plots are based on different treatments of \SC that fail to stop (see text for details)}
    \label{fig:min_cost}
\end{figure}

To determine which criterion is best for a given scenario, we evaluate our cost function $C$ (Eq. \ref{eq:cost}) for a range of values of $m$, $n$ and $l$. However, we first need to specify our treatment of the \SC that fail to stop in a particular experiment. We discuss three approaches: First, we {\em penalize} \SC that do not stop by setting the accuracy and number of labels to the worst values (lowest accuracy and highest number of labels) obtained for that dataset by any criterion. Second, we {\em include} \SC that failed to stop without penalty, ignoring runs on which they fail to stop. Finally, we completely {\em exclude} \SC that fail to stop at least once.

Each of the three approaches is shown in Fig. \ref{fig:min_cost}. The accuracy and number of labels used in Eq. \ref{eq:cost} for each stopping criterion are averages over all runs on all datasets. In regions marked `Indeterminate' (light grey) there is no single significantly best criterion. The x-axis is the cost of the final model misclassifying examples; higher values of $nm$ penalize stopping early before the classifier has reached maximum accuracy. The y-axis is the cost of obtaining labels during training; higher values of $l$ penalize stopping late and using more labels than necessary. If a criterion has a region on these plots, it indicates that it is the optimal choice for a particular range of values of $nm$ and $l$, while \SC that do not have regions on these plots are never the best choice. Penalising methods that fail to stop is a reasonable compromise: when all methods are included, \criteria{Entropy-MCS} appears to perform well but it only stops on a single dataset (see Fig. \ref{fig:failed_to_stop}). Hence, we report the penalized results in subsequent plots.

\begin{figure}
    \centering
    \includegraphics[width=0.8\textwidth]{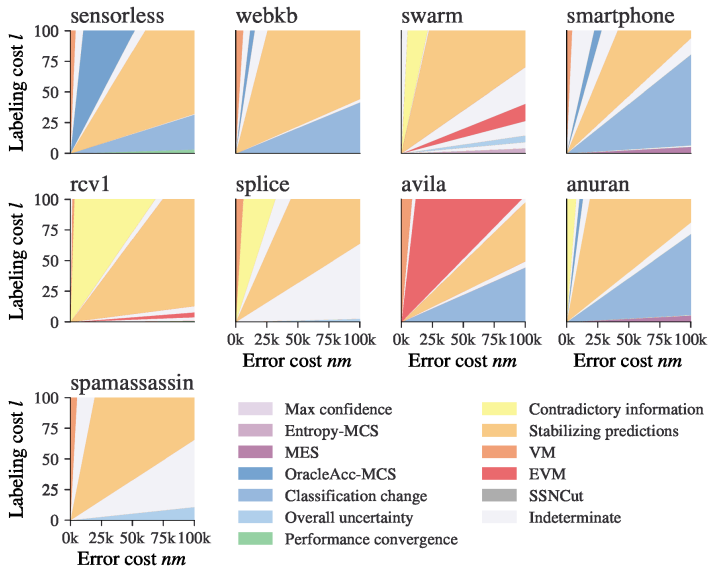}
    \caption{The regions in which each criterion is cost-optimal by dataset, when using SVMs. Criteria were penalised for each run in which they failed to stop}
    \label{fig:min_cost_dataset}
\end{figure}

Figure \ref{fig:min_cost_dataset} demonstrates that there is variation in the \SC's performance across different datasets, particularly in the high and low label cost areas. 
\criteria{Stabilizing predictions} is impressively reliable and consistently dominant in the balanced cost region.
The low label cost region is dominated by \criteria{Classification change} and \criteria{Overall uncertainty}.  \criteria{Classification change}'s performance varies substantially on splice, swarm, and spamassassin, while \criteria{Overall uncertainty} fails to stop entirely on avila and more than half the time on webkb (shown in Fig. \ref{fig:failed_to_stop}).
The early \SC that are preferable in the high label cost region have larger optimal regions on sensorless, rcv1, and avila. This is due to the comparatively poor performance of \criteria{Stabilizing predictions} on these datasets, which is shown in Fig. \ref{fig:pareto_svm}.

The regions of the plots that each \SC dominates (if any) are summarised in Table \ref{tab:criteria_results}. `Cheap labels' corresponds to the lower right section of the plots with a relatively high misclassification cost, `Balanced costs' to the center, and `Expensive labels' to the upper left section.

\subsection{Cost Examples}
\label{sec:experiments:examples}

A key issue with prior evaluations of \SC is that they have not given actionable guidance to practitioners about which to use in a particular use-case \citep{zhu_learning_2008}. Here, we introduce two in-context examples and demonstrate estimating the cost parameters, choosing assumptions, and selecting the best criterion. These are intended as a guide rather than precise evaluations, and practitioners should repeat this process for their own use-case.

\begin{figure}
    \centering
    \includegraphics[width=\columnwidth]{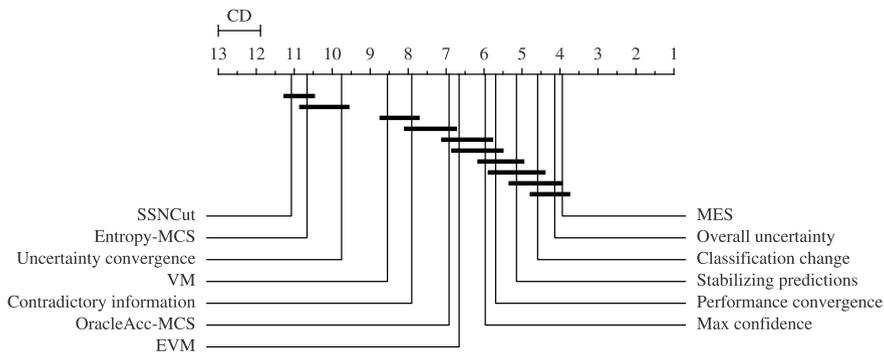}
    \caption{A critical difference diagram of the cost ranking of \SC in the mammogram scenario using an SVM. The horizontal bars group criteria which are not significantly different}
    \label{fig:mamogram_example}
\end{figure}

Breast cancer is a leading cause of death among women worldwide, and there has been interest in using ML to augment the work of radiologists. Because of the training required for radiologists, obtaining labels is expensive and AL is a possible solution.
To find the best criterion in this case we start by estimating the relevant costs. 
We assume the model is in use for one year and classifies $n=336,000$ mammograms in that time period\footnote{The number of screening mammograms performed in six U.S. areas in 2019; \url{breastcancer.org/research-news/mammogram-rates-rebounding-after-pandemic}; accessed on 2021-09-19}, false positives (FP) have no cost (they are subsequently given a correct diagnosis by a radiologist), all initial diagnoses occur at stage 0, and a false negative (FN) is later correctly diagnosed in stage I/II. 
Additionally, we assume FN and FP are equally likely. In practice, this could be estimated by running AL for a few rounds and using the class distribution of labelled instances as the FN/FP distribution.
The median difference in insurance payouts in the U.S. between stage I/II diagnoses and stage 0 diagnoses is $\$21,484$ \citep{blumen_comparison_2016}. Hence, the misclassification cost is $m=\$10,742$.
Our final parameter is the label cost, which we estimate based on the average time it takes for a radiologist to label a mammogram ($4$ minutes), the median salary of radiologists in the U.S. ($\$425,890$), and the number of working hours in a year ($2,087$) \citep{haygood_timed_2009}. Our cost per label is then $l=\$13.60$.
Figure \ref{fig:mamogram_example} presents a critical difference diagram of the ranking of \SC in this scenario. The best criteria are closer to 1, and the horizontal bars group criteria which are not significantly different from each other.
Note that these figures use Friedman and Nemenyi tests so the ranking differs from the cost region plots. 
The best \SC are \criteria{MES}, \criteria{Overall uncertainty}, and \criteria{Classification change} with \criteria{Stabilizing predictions} not far behind. This is a high misclassification cost scenario, so the \SC that stop as late and obtain high accuracy are preferred.

\begin{figure}
    \centering
    \includegraphics[width=\columnwidth]{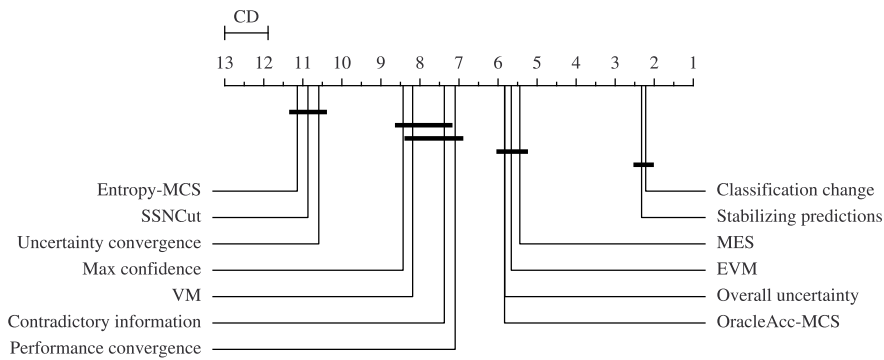}
    \caption{A critical difference diagram of the cost ranking of \SC in the marketing scenario using an SVM. The horizontal bars group criteria which are not significantly different}
    \label{fig:marketing_example}
\end{figure}

Next, we consider an advertiser who aims to send advertisements to potential customers who are most likely to purchase their product. In this case, both false positives and false negatives result in the loss of a sale, and hence the cost of misclassification is this opportunity cost which we set at $m=\$20$. We assume we send a flyer once a week, and that we classify $100$ people each week to determine the best candidates. With a model lifetime of $20$ weeks, we expect to perform $n=2,000$ classifications. In this example, the cost of a label is the cost of surveying a customer to determine whether they would purchase the product, which we estimate at $l=\$1$. This scenario has a relatively small number of lifetime predictions, and we see in Fig. \ref{fig:marketing_example} that the aggressive methods \criteria{Classification change} and \criteria{Stabilizing predictions} are preferred.


\subsection{Neural Networks and Random Forests}
\label{sec:experiments:models}

The behaviour of \SC depends on the learning algorithm. In this section, we describe the most salient differences in \SC performance when using NNs and random forests. The figures are included in the supplementary material.

Under random forests, the most notable difference from SVMs is the stronger performance of \criteria{OracleAcc-MCS}. This \SC has a dominant position in the lower label cost region (using penalty), while it was absent under SVMs. The best \SC from our SVM evaluation, \criteria{Stabilizing predictions} and \criteria{Classification change}, remain dominant in the balanced and high misclassification cost regions and retain top positions in our example cases.

Under NNs, \criteria{OracleAcc-MCS} takes the place of \criteria{Stabilizing predictions} as the \SC which dominates the balanced region in the penalty case. This is partly due to \criteria{Stabilizing predictions} failing to terminate on the avila dataset. 
\criteria{Performance convergence} joined the best group of \SC in our mammogram example, performing substantially better than on SVMs. In the marketing example, \criteria{Stabilizing predictions}'s failure to stop on the avila dataset left \criteria{Classification change} as the best choice.
The correlation between \SC's metrics and the classifier's accuracy also changed substantially; all except \criteria{Contradictory information} and \criteria{Max confidence} had significantly increased correlation coefficients. This is surprising given that most \SC we evaluate were developed on SVMs. However, identifying the cause of this is beyond the scope of this paper.


\raggedbottom
 
\section{Limitations}
\label{sec:limitations}


Our evaluation of \SC found that many of the previously proposed \SC perform inconsistently, or stop too early to produce useful models. This section suggests four possible improvements that could be made to \SC to enhance their performance, or to make their development and evaluation more reliable.


First, we suggest that researchers developing \SC evaluate them on a variety of datasets. Many of the \SC evaluated in this paper were developed and tested on only a single, or a small number of, dataset(s). This is evident in the poor generalisation behaviour of these \SC. Criteria which perform well in specific conditions are only useful if it is possible for a practitioner to determine whether their problem meets these conditions without collecting surplus labels. To simplify development and evaluation of \SC we provide our framework for performing AL, testing \SC, and generating summary figures.


Second, it is clear that while many of the proposed \SC's metrics correlate with the classifier's performance, the conditions used to determine the stopping point are less reliable. 
This is why we suggest particular attention should be given to the condition.
One possible direction could be to train a model to use as the condition. Each of the previously proposed metrics could be used as a feature with the output determining whether to stop on a particular iteration. A complicating factor is the computational cost of collecting enough data to create a reliable condition. Many of the AL runs used to evaluate \SC in this paper took over 24 hours, with some datasets taking over a week. This can be alleviated by using models which support online learning, but even then it may be difficult to obtain enough examples. Similarly, collecting and pre-processing enough datasets to train a generalisable condition would be a challenge.


Instead of learning a condition, an interesting direction is to utilize derivatives of the metric in the condition. Some of the conditions considered in this paper already do this, but with finite differences evaluated over a sliding window. Instead, we suggest learning from dynamical systems modelling and using a regularised numerical derivative such as \textit{total variation regularisation} \citep{chartrand_numerical_2011}. Again, this approach has a complicating factor: we do not know the metric ahead of time, so the condition must be evaluated on the last value, for which less information is available (one neighbour instead of two).


There are additional desirable properties of \SC beyond the number of instances and accuracy at the point they stop. In particular, the ability to tune how early a criterion stops has been discussed as a useful attribute \citep{bloodgood_method_2009}. Indeed, all of the methods we evaluated have at least one hyper-parameter capable of tuning how early the condition stops. Yet, none of these \SC have obvious methods to tune these parameters. Authors most commonly suggested a single value, or in some cases a few values. \citeauthor{dimitrakakis_cost-minimising_2008}'s approach of integrating a cost measure into the criterion to expose understandable hyperparameters is a good one and furthering this approach is a promising line of future research.

\section{Conclusion}
\label{sec:conclusion}




In this paper, we perform the first large comparison of proposed \SC and propose an intuitive cost measure to evaluate the strengths and weaknesses of each criterion, giving practitioners the necessary guidance to implement AL. We compare 13 \SC across nine datasets and three ML models. Our cost measure connects the trade-off between labelling effort and accuracy to three intuitive cost parameters: label cost, model lifetime, and misclassification cost. 
We find that while the best criterion depends on the above parameters, two \SC stand out: \criteria{Stabilizing predictions} reliably stops close to the point at which the classifier's performance saturates, while \criteria{Classification change} reliably stops later in the AL process.


Alongside our comparison, we provide the first public implementations of all \SC we evaluate, making it easy for practitioners to utilize AL to train ML models at a lower cost, or in cases where they were previously infeasible. Additionally, we provide our evaluation framework so that researchers have a baseline upon which to develop new \SC, and suggest promising directions for future work to advance the field of AL.


We believe that by using our cost measure to perform a large-scale comparison, we make AL a sensible choice for practitioners who need to use ML in label-constrained environments. Our results expand the number of situations in which ML can be deployed, its benefits exploited, and ultimately show how the prodigal use of data prevalent in the field today can be mitigated with intentional selection.



\section*{Declarations}
{\small
\textbf{Funding.} The authors did not receive funding from any organization.

\noindent\textbf{Conflicts of Interest.} We do not declare any conflicts.

\noindent\textbf{Ethics.} Not applicable.

\noindent\textbf{Consent to participate.} Not applicable.

\noindent\textbf{Consent for publication.} Not applicable.

\noindent\textbf{Data.} All datasets are publicly available and referenced in the paper.

\noindent\textbf{Code.} 
Available at \href{github.com/zacps/al-stopping-conditions}{https://github.com/zacps/al-stopping-conditions}.

\noindent\textbf{Contributions.} Pullar-Strecker designed and wrote experiments, wrote initial draft. Dost contributed to figures, provided suggestions, edited draft. Wicker and Frank provided suggestions and edited draft.
}

{\footnotesize
\bibliography{References_abbreviated}}

\end{document}